\documentclass[letterpaper]{article} 
\usepackage{aaai2027}  
\usepackage[hyphens]{url}  
\usepackage{graphicx} 
\usepackage{natbib}  
\usepackage{caption} 
\usepackage{algorithm}
\usepackage{algorithmic}
\usepackage{tikz}
\usepackage{graphicx, verbatim}
\usepackage{booktabs}
\usetikzlibrary{calc, positioning, matrix}
\usetikzlibrary{decorations.pathreplacing}
\usetikzlibrary{shapes.geometric, arrows.meta}
\usepackage{twemojis}
\usepackage{pifont}
\usepackage{amsmath}
\usepackage{amssymb}
\usepackage{dsfont}

\usepackage{newfloat}
\usepackage{listings}
\DeclareCaptionStyle{ruled}{labelfont=normalfont,labelsep=colon,strut=off} 
\floatstyle{ruled}
\newfloat{listing}{tb}{lst}{}
\floatname{listing}{Listing}

\usepackage{booktabs}

\title{Enhancing Visual Reasoning in Chest X-Ray Report Generation Using Reinforcement Learning}
\author{
    Written by AAAI Press Staff\textsuperscript{\rm 1}\thanks{With help from the AAAI Publications Committee.}\\
    AAAI Style Contributions by Peter Patel Schneider,
    Sunil Issar,\\
    J. Scott Penberthy,
    George Ferguson,
    Hans Guesgen,
    Francisco Cruz\equalcontrib\corresponding,
    Marc Pujol-Gonzalez\equalcontrib\corresponding
}
\affiliations{
    \textsuperscript{\rm 1}Association for the Advancement of Artificial Intelligence\\

    1101 Pennsylvania Ave, NW Suite 300\\
    Washington, DC 20004 USA\\
    proceedings-questions@aaai.org
}

\title{Enhancing Visual Reasoning in Chest X-Ray Report Generation Using Reinforcement Learning}
\author {
    Denis Musinguzi\textsuperscript{\rm 1}\corresponding,
    Andrew Katumba\textsuperscript{\rm 2},
    Prasenjit Mitra\textsuperscript{\rm 1}
}
\affiliations {
    \textsuperscript{\rm 1}Carnegie Mellon University\\
    \textsuperscript{\rm 2}Makerere University\\
    dmusingu@andrew.cmu.edu, andrew.katumba@mak.ac.ug, pmitra@andrew.cmu.edu
}

\begin{document}

\maketitle

\begin{abstract}
Medical report generation has made significant progress with the rise of modern vision–language models and the growing availability of large‑scale medical datasets. However, hallucinations remain a major challenge, largely due to the limitations of supervised fine‑tuning (SFT), which prioritizes lexical similarity to reference reports rather than clinical correctness. While reinforcement learning has shown strong performance in domains with verifiable rewards such as mathematics and code generation, its application to open‑ended medical tasks remains limited. Existing work focuses on evaluating final answers, overlooking the model’s reasoning, despite evidence that flawed reasoning can degrade overall performance. In this study, we propose a framework that verifies the model’s reasoning process by integrating anatomical regions, bounding boxes, and region‑level textual descriptions. We design spatial and factual reward mechanisms to ensure that the model’s reasoning is both visually grounded and factually accurate. Starting from Qwen3‑VL‑8B‑Instruct as our base model, we adapt it to the medical domain using supervised fine‑tuning, introduce reasoning capability through a cold‑start SFT stage, and refine it with reinforcement learning. We find that RL provides performance gains beyond those achievable through SFT alone, and that jointly verifying both reasoning steps and final outputs yields larger improvements than verifying either in isolation. We further identify multiple modes of reward hacking in the RL stage. Finally, the model’s structured think‑traces enhance interpretability, making its outputs easier to audit for clinical use.
\end{abstract}

\section{Introduction}
\label{sec:intro}

\begin{figure*}[ht]
\centering
\begin{tikzpicture}[
    scale=0.6, 
    transform shape,
    node distance=0.8cm and 1cm,
    box/.style={rectangle, draw, rounded corners=2pt, minimum width=2cm, minimum height=1.5cm, align=center, line width=0.8pt, inner ysep=2pt},
    arrow/.style={-Stealth, line width=0.5pt},
    label_style/.style={font=\sffamily\scriptsize\bfseries}
]

    \begin{scope}[shift={(3.5, 0)}, local bounding box=stage1]
        \node[box] (s1_input) {X-ray \quad \includegraphics[width=0.5cm]{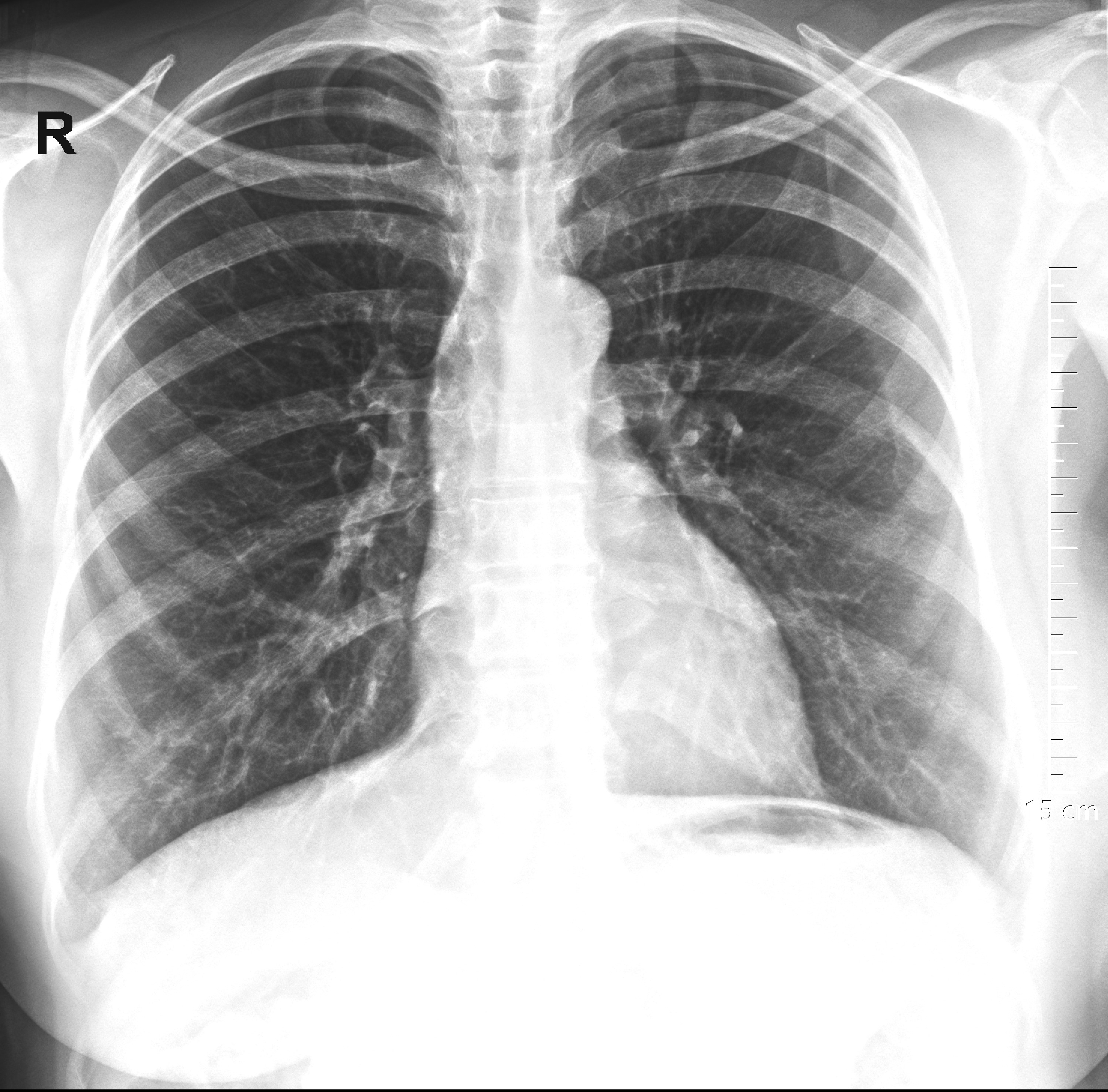} \\ \vspace{0.2cm} {\scriptsize Prompt: Generate report}};
        \node[box, fill=green!5, right=of s1_input] (s1_mllm) {MLLM \quad \scalebox{1.5}{\twemoji{fire}}};
        \node[box, fill=orange!40, right=of s1_mllm] (s1_output) {{\scriptsize \texttt{<answer>...</answer>}
        }};
        
        \draw[arrow] ($(s1_input.east)+(0,0.3)$) -- ($(s1_mllm.west)+(0,0.3)$);
        \draw[arrow] ($(s1_input.east)+(0,-0.3)$) -- ($(s1_mllm.west)+(0,-0.3)$);
        \draw[arrow] (s1_mllm) -- (s1_output);
        
        \node[anchor=west, font=\sffamily\small\bfseries] at ($(s1_input.west)+(0,1.2)$) {Stage 1:  Supervised Finetuning};
    \end{scope}

    \begin{scope}[shift={(3.5,-3.5)}, local bounding box=stage2]
        \node[box] (s2_input) {X-ray \quad \includegraphics[width=0.5cm]{Figures/1.jpg} \\ \vspace{0.2cm} {\scriptsize Prompt: Generate report}};
        \node[box, fill=green!5, right=of s2_input] (s2_mllm) {MLLM \quad \scalebox{1.5}{\twemoji{fire}}};
        \node[box, fill=orange!40, right=of s2_mllm] (s2_output) {{\scriptsize \texttt{<think>...</think>}}\\{\scriptsize \texttt{<answer>...</answer>}}};
        
        \draw[arrow] ($(s2_input.east)+(0,0.3)$) -- ($(s2_mllm.west)+(0,0.3)$);
        \draw[arrow] ($(s2_input.east)+(0,-0.3)$) -- ($(s2_mllm.west)+(0,-0.3)$);
        \draw[arrow] (s2_mllm) -- (s2_output);

        \node[anchor=west, font=\sffamily\small\bfseries] at ($(s2_input.west)+(0,1.2)$) {Stage 2: Cold-start SFT};
    \end{scope}

    \begin{scope}[shift={(0,-7.5)}, local bounding box=stage3]
        \node[box] (s3_input) {X-ray \quad \includegraphics[width=0.5cm]{Figures/1.jpg} \\ \vspace{0.2cm} {\scriptsize Prompt: Generate report}};
        \node[box, fill=green!5, right=of s3_input] (s3_mllm) {MLLM \quad \scalebox{1.5}{\twemoji{fire}}};
        \node[box, fill=orange!40, right=of s3_mllm, minimum height=2cm] (s3_output) {output 1\\output 2\\...\\output n};
        
        \node[box, fill=green!20, minimum height=0.7cm, anchor=north west] (s3_ref) at ([xshift=0.8cm]s3_output.north east) {{\scriptsize ref model}};
        \node[box, fill=green!20, minimum height=0.7cm, below=0.4cm of s3_ref] (s3_rew) {{\scriptsize reward function}};
        
        \node[box, right=3.5cm of s3_output, minimum height=2cm] (s3_rs) {reward 1\\reward 2\\...\\reward n};
        \node[box, right=2.2cm of s3_rs, minimum height=2cm] (s3_adv) {advantage 1\\advantage 2\\ ...\\advantage n};

        \draw[arrow] ($(s3_input.east)+(0,0.3)$) -- ($(s3_mllm.west)+(0,0.3)$);
        \draw[arrow] ($(s3_input.east)+(0,-0.3)$) -- ($(s3_mllm.west)+(0,-0.3)$);
        \draw[arrow] (s3_mllm) -- (s3_output);
        \draw[arrow] (s3_output.east |- s3_ref) -- (s3_ref.west);
        \draw[arrow] (s3_output.east |- s3_rew) -- (s3_rew.west);
        \draw[arrow] (s3_rew.east) -- (s3_rs.west |- s3_rew.east);
        \draw[arrow] (s3_rs) -- node[above, font=\scriptsize] {Normalization} (s3_adv);
        
        \draw[arrow] (s3_ref.north) -- ++(0,0.4) -| node[pos=0.25, above, font=\scriptsize] {KL Divergence} (s3_mllm.north);
        \draw[arrow] (s3_adv.north) -- ++(0,1.2) -| ([xshift=-0.2cm]s3_mllm.north);

        \node[anchor=west, font=\sffamily\small\bfseries] at ($(s3_input.west)+(0,1.5)$) {Stage 3: RL (GRPO)};
    \end{scope}

\end{tikzpicture}
\caption{An overview of the training methodology comprising three stages: (1) supervised fine‑tuning (SFT), during which the model is trained to produce the final report; (2) cold‑start SFT, in which explicit reasoning is incorporated before report generation; and (3) reinforcement learning, wherein the model’s reasoning capabilities are further refined using GRPO.}
\label{fig:training}
\end{figure*}

Medical imaging plays a crucial role in disease diagnosis and screening, particularly for thoracic infections such as tuberculosis (TB) and pneumonia. Chest X‑rays are the most common form of medical imaging \cite{irede2026medical}, but their interpretation requires trained radiologists. The growing demand for chest X‑ray examinations has created an unsustainable workload for the limited radiology workforce~\cite{tanno2025collaboration}, leading to delays in diagnosis, late detection of advanced disease, and high rates of preventable mortality, particularly in low-income countries.

Medical report generation offers a promising way to alleviate this burden. However, its adoption in clinical settings remains limited because current models often fail to meet the required performance standards \cite{tanno2025collaboration} and lack the interpretability that radiologists require. A major challenge is hallucination, where the model generates incorrect findings or omits critical abnormalities present in the image \cite{yang2025mitigating, heiman2025factchexcker}. In clinical practice, such errors can lead to serious misdiagnoses. This behavior stems largely from the standard training objective: models are trained with next-token prediction, which rewards lexical similarity to reference reports rather than grounding descriptions in the image content \cite{bai2024hallucination}.

Reinforcement learning (RL) has become a central paradigm for post-training large language models~\cite{ouyang2022training}. Recent works have shown that combining Reinforcement Learning with Verifiable Rewards (RLVR) can substantially improve performance on mathematics and code generation tasks~\cite{shao2024deepseekmath, guo2025deepseek}. Instead of relying solely on supervised finetuning, these works employ RL algorithms such as Group Relative Policy Optimization (GRPO) \cite{shao2024deepseekmath} and Dynamic Sampling Policy Optimization (DAPO)~\cite{yu2025dapo} to reward model outputs. Building on this progress, several studies have applied RL to closed‑ended medical tasks, including visual question answering and referring expression comprehension~\cite{Liu_2025, shen2025vlm, lai2026med}. However, these approaches focus on closed-ended tasks with short, fixed-response answers, leaving open whether RL can improve the free-form descriptive report generation that clinical practice demands.

Although some recent studies~\cite{liu2026scaling, gundersen2025enhancing} \textbf{apply RLVR to medical report generation, they verify only the model's final output, leaving the reasoning process that produced it unchecked}. This allows models to reach correct conclusions through faulty or clinically implausible reasoning, and, more often, to reach incorrect ones since flawed reasoning degrades the quality of the generated report. Unverified reasoning therefore harms not only interpretability and reliability, both essential for safe clinical deployment, but downstream performance itself.

To address these limitations, we develop rewards that verify both the model's reasoning process and its final output. The reasoning is checked by two rewards that ensure it is visually grounded: a spatial reward, computed as the mean intersection over union across all regions referenced in the reasoning trace, and a factual reward, computed with RadGraph~\cite{jain2021radgraph}. We construct the reasoning data by extracting anatomical regions, their bounding box coordinates, and corresponding regional descriptions from Chest ImaGenome~\cite{wu2021chest}, which provides scene graphs for MIMIC-CXR reports~\cite{johnson2019mimiccxr}, with the final reports taken directly from MIMIC-CXR. Starting from a general-purpose Qwen3‑VL‑8B‑Instruct~\cite{bai2025qwen3vltechnicalreport} checkpoint, we adapt the model to the medical domain through a three-stage pipeline illustrated in Figure~\ref{fig:training}. We find that verifying both the reasoning and the final output yields better performance than verifying either alone.
In summary, our contributions are as follows:
\begin{enumerate}
    \item \textbf{A gated reward combination} in which factual verification of a regional finding is conditional on that region being correctly localized rather than an additive combination.
    \item We \textbf{combine process-based and final answer verification}, supervising region localization and findings alongside the radiology report. 
    \item We identify \textbf{two reward-hacking modes from single rewards}: spatial-only rewards yield well-localized but duplicated findings, and factual-only rewards yield correct findings on wrong regions, scoring below no reward at all.
\end{enumerate}

\section{Related Work}

\subsection{Multimodal Reasoning}
Explicit reasoning and verifiable rewards have enabled RL in Large Language Models (LLMs) beyond traditional reinforcement learning from human feedback (RLHF). Reasoning models such as DeepSeek-R1~\cite{shao2024deepseekmath} and Qwen3~\cite{yang2025qwen3} combine cold‑start supervised finetuning, which introduces explicit reasoning steps, with RL to further enhance the model’s reasoning capabilities. RLVR~\cite{guo2025deepseek} has produced substantial performance gains in tasks such as mathematics and code generation.

Inspired by the success of explicit reasoning in LLMs, several works have incorporated similar mechanisms into vision-language models. Vision-R1~\cite{huang2025vision} generates multimodal chain-of-thought data and applies GRPO to Qwen2.5-VL, while Visual-RFT~\cite{liu2025visual} and VLM-R1~\cite{shen2025vlm} report improved localization and visual understanding over standard SFT with limited additional data. These methods, however, target closed-ended tasks whose outputs—a math answer, a class label, or a bounding box—can be matched exactly with ground truth. We depart from this setting by applying RL to report generation, an open-ended task whose free-form outputs admit no single correct answer and are substantially harder to verify.

\subsection{VLMs in Radiology}
There has been tremendous progress in chest X‑ray report generation due to advancements in VLMs trained with supervised finetuning (SFT) on large‑scale datasets such as MIMIC‑CXR~\cite{johnson2019mimic} and CheXpert‑Plus~\cite{chambon2024chexpert}. Models including MedGemma~\cite{sellergren2025medgemma}, LLaVA‑Med~\cite{li2023llava}, and MAIRA~\cite{bannur2024maira2groundedradiologyreport} generate increasingly plausible reports. 

More recently, the success of RL in LLMs and VLMs has inspired applications in medical imaging. Med-R1 \cite{lai2026med}, MedVLM-R1 \cite{pan2025medvlm}, and ChestX-Reasoner \cite{fan2025chestx} apply GRPO to multiple-choice visual question answering, achieving high accuracy and structured reasoning. MedGround-R1 \cite{xu2025medground} and DeepMedix-R1 \cite{lin2025foundation} extend this direction to medical image grounding using spatial and factual rewards, while RadVLM-GRPO \cite{gundersen2025enhancing} applies rewards to report generation and visual grounding for state-of-the-art performance. GEMeX-CoT \cite{Liu_2025} and UniRG-CXR \cite{liu2026scaling} instead optimize natural-language metrics directly through RL rather than relying on supervised fine-tuning alone. In all of these works, spatial and factual rewards are applied in isolation. We differ by combining them in a gated mechanism that jointly verifies the model's reasoning process.

Notably, explicit reasoning is not always beneficial: models that reason before answering can underperform those that answer directly \cite{li2025think}. We argue that this occurs when the reasoning itself is left unverified—free to be ungrounded or clinically implausible even when the final answer is correct. Prior methods verify only the final output, leaving this reasoning unchecked. In this work, we instead make the reasoning process itself verifiable: we jointly apply spatial and factual rewards to the reasoning, rewarding the model for correctly identifying and describing anatomical regions before synthesizing the final narrative. This ensures that the model's path to the answer is as verifiable as the answer itself.

\section{Method}
The overall training pipeline is shown in Figure \ref{fig:training}. We train our model in three stages to progressively enhance its domain knowledge, reasoning ability, and visual grounding. The stages include: (1) Supervised Finetuning (SFT), (2) Cold‑Start SFT for reasoning induction, and (3) RL with  GRPO. 

We start with a general-purpose vision-language Qwen3-VL-8B-Instruct~\cite{bai2025qwen3vltechnicalreport} model and train it using SFT to adapt it to the medical domain. We train the model to generate the final report directly using standard next-token prediction for vision language models by maximizing the conditional likelihood:
\begin{equation}
    \label{eqn:sft}
    \mathcal{L}_{SFT} = -\sum_{i=1}^{L} \text{log}~p(x_i|x_v, x_{q}, x_{a<i})
\end{equation}
where $x_v$ denotes the visual tokens, $x_{q}$ and $x_{a<i}$ represent the prompt and response preceding token $x_i$ respectively. $L$ is the number of tokens in the target sequence. 

Starting from the checkpoint obtained after the SFT stage, we introduce explicit reasoning into the model using structured reasoning data. This data includes reasoning traces enclosed in \texttt{<think>...</think>} tags and the final answer enclosed in \texttt{<answer>...</answer>} tags. Each reasoning trace consists of anatomical region names, bounding‑box coordinates, and the corresponding findings for each region in the image, all organized into a dictionary‑based structure. This format encourages the model to “think before answering”. The combination of bounding boxes and regional descriptions forces the model to ground its reasoning in the chest X-ray image, thereby reducing the likelihood of generating incorrect findings. The final report is a result of synthesizing region‑level descriptions.

Following the cold‑start SFT, we train the model using RL with the GRPO algorithm. This stage is designed to shift the model’s objective from simple next-token prediction to systematic diagnostic reasoning.
The model is incentivized to adopt a hierarchical, ``bottom-up'' workflow. It is rewarded for successfully decomposing the global image into specific anatomical regions, performing localized assessments, and only then synthesizing these observations into a holistic radiology report. By enforcing this region-by-region examination, the framework aligns the model's internal reasoning with the actual visual attributes of the chest X-ray.

This structured scrutiny serves as a critical defense against hallucinations. In traditional generative models, hallucinations often arise from a ``top-down'' bias where the model generates frequent phrases like ``no pneumothorax'' without verifying the underlying pixels. By requiring the model to generate verifiable spatial bounding boxes and regional descriptions during the reasoning process, our RL strategy ensures that every diagnostic claim in the final report is anchored in localized visual evidence. Consequently, the model evolves from a linguistic imitator into a grounded diagnostic agent.

At each RL training step, we sample $G$ candidate completions $\{o_{1},  o_2,...,o_G\}$ from an old policy, $\pi_{old}$. Each sampled completion receives a reward $r_i$. The rewards are used to compute the relative advantage $A_{i,t}$:
\begin{equation}
\label{eqn:adv}
    A_{i,t} = \frac{r_i - \text{mean}(\{r_1, r_2,...,r_G\})}{\text{std}(\{r_1, r_2,...,r_G\})}
\end{equation}
which normalizes the aggregated rewards across sampled completions. This ensures that completions outperforming the group baseline receive a positive advantage, while underperforming samples receive a negative advantage. 
The advantage is used to update the on-policy model, $\pi_{\theta}$, according to the GRPO objective in equation \ref{eqn:grpo}.

\begin{equation}
\label{eqn:grpo}
\begin{split}
\mathcal{J}_{GRPO}(\theta) &= \mathbb{E}_{(q,a) \sim \mathcal{D},\, \{o_i\}_{i=1}^{G} \sim \pi_{old}(\cdot|q)}
\Biggl[ \frac{1}{G} \sum_{i=1}^{G} \frac{1}{|o_i|} \sum_{t=1}^{|o_i|} \\
&\hspace{-4em} A_{i,t} \Bigl( \min \bigl( r_{i,t}(\theta),\, \mathrm{clip}( r_{i,t}(\theta),\, 1-\epsilon_{low},\, 1+\epsilon_{high} ) \bigr) \\
&\quad - \beta D_{KL}(\pi_{\theta} \| \pi_{ref}) \Bigr) \Biggr]
\end{split}
\end{equation}

\begin{equation}
\label{eqn:reward}
    r_{i,t} (\theta) = \frac{\pi_{\theta}(o_{i,t}|q, o_{i<t})}{\pi_{\theta_{old}}(o_{i,t}|q, o_{i<t})}
\end{equation}

where $\beta$, $\epsilon_{low}$ and $\epsilon_{high}$  are hyper-parameters. The KL-divergence term $\mathcal{D}_{KL}$, regularizes the model, preventing $\pi_{\theta}$ from drifting too far from the reference model $\pi_{ref}$. We use the unbiased estimator defined by \cite{schulman2020kl,shao2024deepseekmath}.

We verify the model's reasoning through a cascade of three checks in which each stage gates the next: structural format gates spatial localization, and spatial localization gates factual verification. Format is treated as a precondition rather than a reward term because it is easily satisfied and, as an additive term, overshadows the signal from the other two.

\textbf{Format}. The reasoning process must be enclosed in \texttt{<think> … </think>} tags and the final report within \texttt{<answer> … </answer>} tags, and the reasoning trace must contain bounding box coordinates \texttt{[$x_{min}, y_{min}, x_{max}, y_{max}$]} for every anatomical region. We verify this structure with regular expressions. Completions that fail receive no reward and are not evaluated further.

\textbf{Spatial Reward}. For completions that clear the format check, the spatial reward evaluates how well the model localizes anatomical regions in the chest X-ray. We extract the predicted regions, match them to the ground-truth regions by anatomical name, and compute the intersection-over-union (IoU) for each matched pair. Regions present in the ground truth but absent from the generated output receive a spatial reward of 0. The spatial reward is the mean IoU across all ground-truth regions.

\textbf{Factual reward}. The factual reward evaluates how accurately the model describes regional findings, comparing the predicted findings against the ground truth for each anatomical region. We compute it with RadGraph-F1, using RadGraph-XL as the RadGraph model. Rather than adding this term unconditionally, we gate it on spatial correctness, giving the total reward:
\begin{equation}
\label{eqn:reward}
    R=R_{\text{spatial}}+ \mathds{1} [\text{IoU}>\tau]R_{\text{factual}}, \tau=0.25
\end{equation}
Regions whose IoU falls below $\tau$ contribute only their spatial term to the total reward.
\section{Experiments}
\subsection{Datasets}
We use two datasets for experiments: the MIMIC-CXR database \cite{johnson2019mimiccxr} and the Chest ImaGenome dataset \cite{wu2021chestimagenome}. The MIMIC-CXR database consists of 377,110 chest X-ray images associated with 227,835 radiology reports. We filter out lateral views and reports without ``Findings'' and ``Impressions'' sections, resulting in a dataset of 213,073 image-report pairs across the official train, validation and test splits. We remove all references to prior studies from the reports, since such text is not supported by a single image.

To facilitate the model's transition from standard generation to verifiable reasoning, we leverage the granular annotations in the Chest ImaGenome dataset. It provides scene graphs for the MIMIC-CXR corpus that map clinical phrases to anatomical regions in the chest. It consists of 642,703 bounding box annotations for 38 anatomical regions together with phrases describing them. We utilize these mappings to construct ``cold-start'' reasoning traces. For each anatomical region, we generate a structured dictionary containing the region name, its bounding box, and a phrase describing its clinical findings in the following format: \textit{\{`\textbf{region}': anatomical region name, `\textbf{bbox}': \texttt{[$x_{min}, y_{min}, x_{max}, y_{max}$]}, `\textbf{findings}': description of the region\}}. The final training samples are organized into a unified, reasoning-centric format designed to distinguish between the visual reasoning process and the final report. The set of anatomical dictionaries constitutes the reasoning trace, which is enclosed in \texttt{<think>} tags. The comprehensive report follows, enclosed in \texttt{<answer>} tags. This structure effectively mirrors the systematic ``region-by-region'' examination used by professional radiologists. Training the model on this format incentivizes it to develop a spatial-factual correspondence that makes its outputs verifiable and traceable to specific visual evidence.

\begin{table*}[t]
\centering
\caption{Results of our model and other baselines on the MIMIC-CXR dataset test set. It includes the results after each training stage. 
\textbf{Bold} values indicate the best performance, while \underline{underlined} values indicate the model with the second-best performance.
}\label{tab:results}
{
\begin{tabular}{l|c|c|c|c|c}
\hline
Model & BLEU-2 & ROUGE-L & METEOR & RADGRAPH-F1 & BERTSCORE\\
\hline
Qwen3-VL-8B-Instruct\cite{bai2025qwen3vltechnicalreport}&4.4\small{$\pm$1.10}&14.0\small{$\pm$1.00}&21.3\small{$\pm$1.10}&10.8\small{$\pm$1.50}&42.0\small{$\pm$1.90}\\
MedGemma\cite{sellergren2025medgemma}&9.6\small{$\pm$1.45}&20.7\small{$\pm$0.25}&26.1\small{$\pm$0.35}&15.5\small{$\pm$0.80}&47.7\small{$\pm$1.35}\\
DeepMedix-R1\cite{lin2025foundation}&21.5\small{$\pm$0.56}&22.3\small{$\pm$0.33}&29.0\small{$\pm$0.57}&18.6\small{$\pm$0.22}&52.8\small{$\pm$1.46}\\
MAIRA-2\cite{bannur2024maira}&14.2\small{$\pm$1.00}&17.7\small{$\pm$0.60}&21.5\small{$\pm$0.50}&12.9\small{$\pm$0.30}&46.6\small{$\pm$1.50}\\
UniRG-CXR \cite{liu2026scaling}&\textbf{26.2}\small{$\pm$1.20}&26.4\small{$\pm$0.45}&\underline{31.5}\small{$\pm$0.85}&\underline{26.7}\small{$\pm$0.46}&47.8\small{$\pm$0.75}\\
RadVLM-GRPO\cite{gundersen2025enhancing}&22.5\small{$\pm$0.20}&\textbf{30.0}\small{$\pm$0.10}&30.2\small{$\pm$1.00}&25.7\small{$\pm$0.10}&\textbf{59.0}\small{$\pm$0.70}\\
Qwen3-VL-8B-GRPO (ours) &\underline{23.4}\small{$\pm$0.16}&\underline{28.5}\small{$\pm$0.20}&\textbf{32.5}\small{$\pm$1.47}&\textbf{27.9}\small{$\pm$0.07}&\underline{56.3}\small{$\pm$0.47}\\
\hline
\end{tabular}}
\end{table*}

\begin{table*}[t]
    \caption{Results of various models on the test split of the IU X-ray dataset, considering only frontal chest X-ray images. \textbf{Bold} values indicate the best performance while \underline{underlined} values indicate the second-best performance. }
    \centering
    \begin{tabular}{l|c|c|c|c|c}
    \hline
         Model & BLEU-2 & ROUGE-L & METEOR & RADGRAPH-F1 & BERTSCORE\\
         \hline
         Our model&\textbf{27.51}\small{$\pm$0.39}&\textbf{34.47}\small{$\pm$0.26}&\underline{39.92}\small{$\pm$0.21}&\textbf{36.53}\small{$\pm$0.23}&\textbf{61.49}\small{$\pm$0.05}\\
         MAIRA-2&21.90\small{$\pm$0.34}&27.40\small{$\pm$0.26}&33.10\small{$\pm$0.26}&27.10\small{$\pm$0.24}&47.70\small{$\pm$0.04}\\
         DeepMedix-R1&23.78\small{$\pm$0.37}&28.53\small{$\pm$0.28}&33.16\small{$\pm$0.27}&27.66\small{$\pm$0.23}&52.97\small{$\pm$0.04}\\
         MedGemma&3.24\small{$\pm$0.38}&9.07\small{$\pm$0.42}&22.24\small{$\pm$0.86}&24.18\small{$\pm$0.22}&17.25\small{$\pm$0.08}\\
         RadVLM-GRPO&20.50\small{$\pm$0.48}&27.26\small{$\pm$0.37}&29.40\small{$\pm$0.74}&24.70\small{$\pm$0.35}&53.14\small{$\pm$0.10}\\
         UniRG-CXR&\underline{27.46}\small{$\pm$0.52}&\underline{33.89}\small{$\pm$0.25}&\textbf{40.12}\small{$\pm$1.00}&\underline{36.12}\small{$\pm$0.18}&\underline{60.70}\small{$\pm$0.06}\\
         \hline
    \end{tabular}
    \label{tab:iu-xray-eval}
\end{table*}

\begin{table}[t]
    \centering
    \caption{Bleu-2 (BL-2), Rouge-L (RG-L), RadGraph-F1
(RG-F1), Meteor, and BertScore (BS) results after different stages of the training pipeline. SFT denotes the model after the SFT stage, C-S SFT denotes the model after the cold-start SFT stage and RL denotes the model after the RL stage. \textbf{Bold} values indicate the best performance while \underline{underlined} values indicate the second-best performance.}
    \begin{tabular}{l|c|c|c|c|c}
    \hline
         Stage&BL-2 &RG-L&MR&RG-F1&BS \\
         \hline
         SFT&\underline{22.0}&\underline{25.3}&\underline{27.0}&16.9&52.0\\
         C-S SFT&14.9&23.2&26.1&\underline{20.4}&\underline{54.2}\\
         RL&\textbf{23.4}&\textbf{28.5}&\textbf{32.5}&\textbf{27.9}&\textbf{56.3}\\
         \hline
    \end{tabular}
    \label{tab:stages}
\end{table}

\subsection{Implementation Details} In the \textbf{supervised finetuning (SFT)} stage, we initialize the model from pretrained Qwen3‑VL‑8B‑Instruct\footnote{https://huggingface.co/Qwen/Qwen3-VL-8B-Instruct} weights and adapt it to the medical domain on 120,000 image-report pairs from the MIMIC-CXR dataset. We train for 2 epochs with the negative log-likelihood loss in Equation~\ref{eqn:sft}, using the AdamW optimizer, a learning rate of $1\times10^{-5}$, a cosine annealing scheduler with a warmup ratio of 0.1, and a global batch size of 128. This stage runs on 8 NVIDIA A100 GPUs with 80 GB VRAM and takes 3 hours.

In the \textbf{cold‑start finetuning stage}, we initialize the model with the final checkpoint after the SFT stage and train on 13,000 training examples comprising structured reasoning traces and final reports, introducing explicit reasoning supervision. We train for one epoch with the negative log-likelihood loss in Equation~\ref{eqn:sft}, using the AdamW optimizer, a learning rate of $5\times10^{-6}$, a cosine annealing scheduler with a warmup ratio of 0.1, and a global batch size of 64. This stage runs on 8 NVIDIA A100  GPUs with 80 GB VRAM and takes 1.5 hours. It equips the model to generate structured intermediate reasoning steps before synthesizing the final report.

In the \textbf{RL stage}, we post-train the model on 74,000 examples, each pairing an image with a reasoning trace and report. We optimize the GRPO objective in Equation~\ref{eqn:grpo} using the AdamW optimizer with a learning rate of $1\times10^{-6}$, a warmup ratio of 0.1, and a batch size of 32. For each prompt, we sample eight rollouts with a maximum generation length of 1024 tokens, and apply a KL-divergence penalty with coefficient $\beta=0.04$ to constrain deviations from the reference policy. Following DAPO~\cite{yu2025dapo}, we employ asymmetric clipping with $\epsilon_{low}=0.2$ and $\epsilon_{high}=0.28$, which encourages exploration and prevents entropy collapse by permitting larger policy updates in the direction of positive rewards. This stage runs on 8 NVIDIA A100 GPUs with bfloat16 mixed precision and DeepSpeed ZeRO-3 and takes 35 hours.


\subsection{Baselines}
\label{baselines}
We compare our approach against a diverse set of vision-language models, including a general-purpose model, Qwen3-VL-8B-Instruct \cite{bai2025qwen3vltechnicalreport}, and five domain-specific models: MedGemma \cite{sellergren2025medgemma}, DeepMedix-R1 \cite{lin2025foundation}, MAIRA-2 \cite{bannur2024maira}, RadVLM-GRPO \cite{gundersen2025enhancing}, and UniRG-CXR \cite{liu2026scaling}. Additionally, we report the performance of our model after each training stage to quantify its individual contribution.

We evaluate all the models using the same protocol. We utilize frontal chest X-ray images from the filtered MIMIC-CXR test set and the IU X-ray test set, preprocessing reports to remove references to prior studies. For our model, we extract the content inside \texttt{<answer>} tags. For models with publicly available checkpoints (Qwen3-VL-8B-Instruct, MedGemma, DeepMedix-R1, and MAIRA-2), we perform inference directly on our filtered test split. For UniRG-CXR and RadVLM-GRPO, whose pretrained weights are not publicly available, we reproduce the models by following the training procedures described in their respective publications.

\subsection{Evaluation Metrics} 
We evaluate report quality using standard lexical similarity metrics, including BLEU-2~\cite{papineni-etal-2002-bleu}, ROUGE-L~\cite{lin2004rouge}, METEOR~\cite{banerjee2005meteor}, and BERTScore~\cite{zhang2019bertscore}. 
To assess clinical factuality, we report RadGraph-F1~\cite{jain2021radgraph} using the RadGraph-XL model.
All metrics are computed using the RadEval~\cite{xu2025radeval} framework. We report the mean and standard deviation over three seeds.

\section{Results}
Tables~\ref{tab:results} and~\ref{tab:iu-xray-eval} show the performance of the models on MIMIC-CXR and IU X-ray. On the MIMIC-CXR, the general-purpose vision-language model Qwen3-VL-8B-Instruct performs substantially worse than models developed specifically for radiology report generation, highlighting the importance of domain-specific training. Among the domain-specific models, UniRG-CXR and RadVLM-GRPO are the strongest overall: UniRG-CXR leads on BLEU-2 score and RadVLM-GRPO records the best ROUGE-L and BERTScore.

Our model, Qwen3-VL-8B-GRPO, leads on clinical factuality with the highest RadGraph-F1 (27.9) and attains the best METEOR score (32.5), while remaining competitive on ROUGE-L (28.5), BLEU-2 (23.4), and BERTScore (56.3). This balance across lexical and clinical criteria indicates that the gains in factuality do not come at the cost of fluency.

Figure \ref{tab:stages} shows the performance of the model across different training stages. Supervised fine-tuning (SFT) provides a strong lexical baseline (BLEU-2 22.0) but weak clinical factuality (RadGraph-F1 16.9). The cold-start stage improves factuality to 20.4 while producing shorter reports, which lowers BLEU-2 to 14.9 through the brevity penalty. GRPO recovers this lexical performance (BLEU-2 23.4) and delivers the largest factuality gain of the pipeline (RadGraph-F1 27.9), indicating that RL improves clinical correctness without sacrificing fluency.

On the IU X-ray dataset, our model generalizes well despite being trained exclusively on MIMIC-CXR. It records the best performance on four out of five metrics (BLEU-2, ROUGE-L, RadGraph-F1, and BERTScore), while UniRG-CXR obtains a slightly higher METEOR score. These results demonstrate that the improvements from our training pipeline transfer effectively across datasets and are not limited to the training dataset.

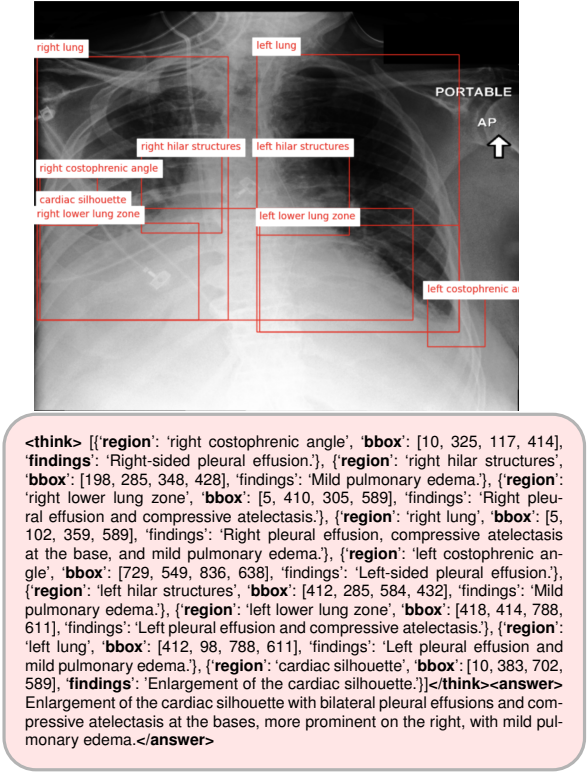
\begin{figure}[ht]
    \centering
    \begin{minipage}{0.45\textwidth}
        \centering
        \includegraphics[width=0.8\textwidth]{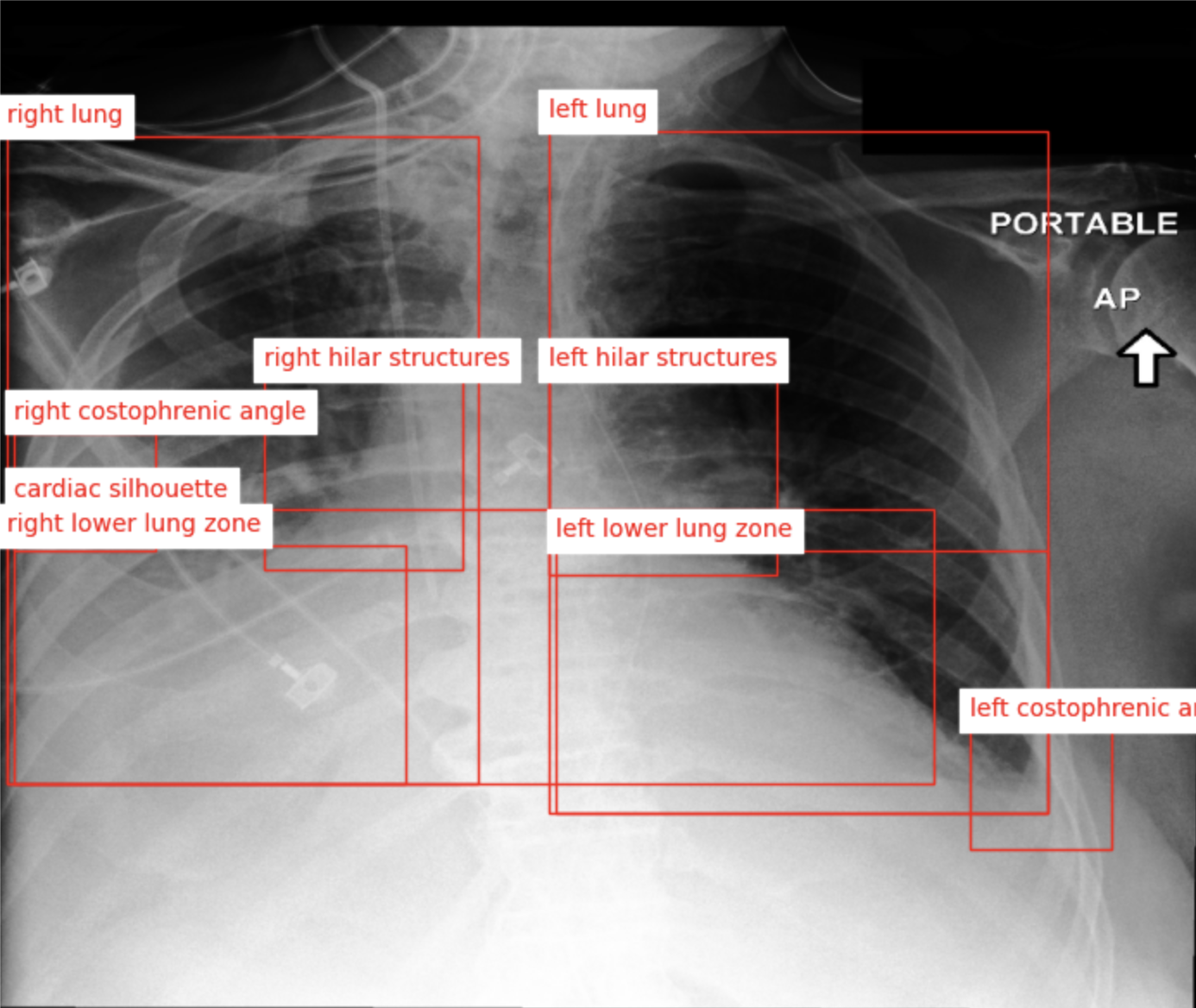}
    \end{minipage}
    \hfill 
    \begin{minipage}{0.5\textwidth}
        \centering
        \begin{tikzpicture}
            \node [
                draw=gray!60, 
                fill=red!10, 
                very thick, 
                rectangle, 
                rounded corners=12pt, 
                inner sep=8pt, 
                text width=0.8\textwidth, 
                font=\tiny\sffamily,
                align=justify,
            ] (textbox) {
            {\tiny \textbf{<think>} [\{`\textbf{region}': `right costophrenic angle', `\textbf{bbox}': [10, 325, 117, 414], `\textbf{findings}': `Right-sided pleural effusion.'\},
            \{`\textbf{region}': `right hilar structures', `\textbf{bbox}': [198, 285, 348, 428], `findings': `Mild pulmonary edema.'\},
            \{`\textbf{region}': `right lower lung zone', `\textbf{bbox}': [5, 410, 305, 589], `findings': `Right pleural effusion and compressive atelectasis.'\},
            \{`\textbf{region}': `right lung', `\textbf{bbox}': [5, 102, 359, 589], `findings': `Right pleural effusion, compressive atelectasis at the base, and mild pulmonary edema.'\},
            \{`\textbf{region}': `left costophrenic angle', `\textbf{bbox}': [729, 549, 836, 638], `findings': `Left-sided pleural effusion.'\},
            \{`\textbf{region}': `left hilar structures', `\textbf{bbox}': [412, 285, 584, 432], `findings': `Mild pulmonary edema.'\},
            \{`\textbf{region}': `left lower lung zone', `\textbf{bbox}': [418, 414, 788, 611], `findings': `Left pleural effusion and compressive atelectasis.'\},
            \{`\textbf{region}': `left lung', `\textbf{bbox}': [412, 98, 788, 611], `findings': `Left pleural effusion and mild pulmonary edema.'\},
            \{`\textbf{region}': `cardiac silhouette', `\textbf{bbox}': [10, 383, 702, 589], `\textbf{findings}': 'Enlargement of the cardiac silhouette.'\}]\textbf{</think><answer>} Enlargement of the cardiac silhouette with bilateral pleural effusions and compressive atelectasis at the bases, more prominent on the right, with mild pulmonary edema.\textbf{</answer>}}
            };
        \end{tikzpicture}
    \end{minipage}
    \caption{\textbf{Qualitative evaluation on the MIMIC-CXR test set.} Predicted region boxes, the corresponding reasoning trace, and the generated report. 
    }
\label{fig:qualitative}
\end{figure}

Figure~\ref{fig:qualitative} illustrates the model's reasoning process. The model traverses the anatomical regions of the image, generating bounding-box coordinates and findings for each before synthesizing them into a final report. The final report is not a concatenation of the trace. Statements such as ``bilateral pleural effusions'' and ``more prominent on the right'' appear in no single region entry but are aggregated from separate left- and right-sided observations, indicating that the model reasons over the regions rather than merely transcribing them. Because every claim remains traceable to specific regions, the outputs are auditable in a way that end-to-end generation does not permit.

\subsection{RL Enhances Fine-Grained Understanding of Chest X-ray Anatomy.}
Training with RL leads to substantial gains in the spatial reward, which tracks the mean IoU across the regions predicted by the model. This rise in spatial reward is accompanied by an improvement in the factual reward.
Notably, when using only spatial rewards, the model exhibits degenerate behavior. It generates identical findings across different regions, despite correctly detecting their locations. This issue can be interpreted as reward hacking, where improvements in localization do not lead to more accurate clinical descriptions. These results highlight the importance of jointly tracking grounding and factual rewards: only their combined optimization prevents reward‑hacking behavior and produces clinically reliable reports that remain firmly grounded in radiological evidence.

\subsection{Ablation Studies}
\begin{table}[t]
\centering
\caption{Results showing Rouge-L (RG-L), Bleu-2 (BL-2), Meteor, RadGraph-F1 (RG-F1) and Bertscore (BS) results for verification of the answer (Ans Verif), the reasoning process (Reas Verif), and the joint verification (Joint Verif).}
\label{tab:ablation1}
{
\begin{tabular}{l|c|c|c|c|c}
\hline
Model &  RG-L & BL-2& Meteor & RG-F1 & BS\\
\hline
Ans Verif&26.2&23.2&23.1&21.0&53.8\\ 
Reas Verif& 25.5&22.1&\textbf{33.5}&23.2&55.2\\
Joint Verif&\textbf{26.8}&\textbf{23.3}&{32.5}&\textbf{27.9}&\textbf{55.6}\\
\hline
\end{tabular}}
\end{table}
\subsubsection{Verification of both the final answer and the reasoning process yields the best performance.} 

Table~\ref{tab:ablation1} compares three reward configurations: (i) rewarding the reasoning process only, (ii) rewarding the final report only, and (iii) jointly rewarding both the reasoning process and the final report. Verification of the reasoning process alone consistently outperforms rewarding only the final report, highlighting the importance of supervising the model's intermediate reasoning steps. By encouraging the generation of clinically coherent step-by-step rationales, the reasoning-based objective provides denser supervision, leading to more effective credit assignment and more stable policy optimization.

Jointly rewarding both the reasoning process and the final report yields the best performance on most metrics. Although reasoning supervision encourages the model to generate accurate regional findings, it does not explicitly enforce how these findings should be summarized into the final report. Consequently, the reasoning-only model often lists findings independently. For example, ``There is a left pleural effusion'' and ``There is a right pleural effusion'' are stated separately instead of producing the clinically appropriate summary, ``There is bilateral pleural effusion.'' This mismatch with the reference reports reduces downstream performance despite the underlying findings being correct. In addition, the reasoning-only model generates slightly longer final reports than the jointly rewarded model, reflecting its tendency to enumerate individual regional findings rather than summarize them into concise clinical statements.

A qualitative analysis of the generated reports further shows that verifying the reasoning process reduces omission errors and decreases hallucinated findings. We attribute these gains to the combined spatial and factual rewards used during training, which jointly constrain the model to produce grounded and semantically consistent outputs.

\subsubsection{Sensitivity Analysis of IoU Gating Threshold for the Factual Reward}
We evaluated the effect of the IoU gating threshold used for the factual reward using thresholds of 0.0, 0.25, and 0.5. To reduce computational cost, we performed RL using 25\% of the training data after completing SFT and cold-start SFT on the full training dataset. The resulting models were then evaluated on the validation set. The results are shown in Table~\ref{tab:thresh}.

Using an IoU threshold of 0.0 consistently reduced downstream performance across all evaluation metrics. Without an IoU constraint, factual rewards could be assigned even when the predicted region did not sufficiently overlap the ground-truth region, weakening the correspondence between localized image content and the generated regional description. In contrast, increasing the threshold to 0.5 produced the lowest overall performance among the evaluated thresholds. A stricter gating criterion reduced the number of training examples that received a positive reward, particularly during the early stages of RL, making optimization less stable.

An intermediate threshold of 0.25 achieved the best overall performance, providing an effective balance between enforcing accurate spatial localization and supplying sufficiently frequent reward signals during training. Based on these results, we use an IoU threshold of 0.25 for all subsequent experiments.

\begin{table}[t]
    \caption{Bleu-4 (BL-4), Rouge-L (RG-L), RadGraph-F1 (RG-F1), Meteor, and BertScore (BS) results of the model at different IoU thresholds for gating the regional factual reward. 
    }
    \centering
    \begin{tabular}{c|c|c|c|c|c}
    \hline
         {IoU Thres.}& {BL-4 }& {RG-L} &{RG-F1}&{Meteor}& {BS} \\
         \hline
         0.00&4.69&25.17&20.27&26.87&54.27\\
         0.25&  \textbf{7.89}&\textbf{27.12}&\textbf{23.73}&\textbf{32.25}&\textbf{55.25}\\
         0.50& 4.98&24.78&20.21&27.62&54.74\\
         \hline
    \end{tabular}
    \label{tab:thresh}
\end{table}

\subsubsection{Spatial and Factual Rewards}
We conducted an ablation study to evaluate the contribution of the spatial and factual reward components during RL. All models were first trained using supervised fine-tuning (SFT) and cold-start SFT on the full training dataset. RL was then performed using 25\% of the training data, after which the models were evaluated on the validation set. We set the IoU threshold to 0.25 for this ablation. Table~\ref{tab:rewards} summarizes the results.

Removing the spatial reward while retaining the factual reward resulted in the poorest overall performance across most evaluation metrics. Without explicit spatial supervision, the model could receive a factual reward even when the generated description was associated with an incorrect image region. This observation is further supported by the fact that the factual-only model performed worse than the model trained without either reward, suggesting that factual feedback alone can reinforce incorrect region-text associations.

In contrast, retaining the spatial reward while removing the factual reward produced performance that was competitive with the full reward formulation. This indicates that spatial supervision plays a dominant role in encouraging the model to associate regional descriptions with the correct anatomical locations.

The best overall performance was achieved when both spatial and factual rewards were used. The spatial reward encourages accurate localization, while the factual reward promotes faithful description generation once the correct region has been identified. Together, the two rewards provide complementary supervision, leading to the strongest downstream performance.

\begin{table}[t]
    \caption{
    Bleu-4 (BL-4), Rouge-L (RG-L), RadGraph-F1 (RG-F1), Meteor, and BertScore (BS) results of the ablation  on factual and spatial rewards. Spat stands for spatial reward and Fac stands for factual reward.
    }
    \centering
    \begin{tabular}{c|c|c|c|c|c|c}
    \hline
        Spat&Fac&BL-4&RG-L&RG-F1&Meteor&BS\\
        \hline
         \ding{55}&\ding{51}&4.69&25.17&20.27&26.87&54.27 \\
         \ding{51}&\ding{55}&6.63&25.27&20.68&28.46&52.30  \\
         \ding{55}&\ding{55}&5.36&24.74&21.18&28.85&55.22  \\
         \ding{51}&\ding{51}&\textbf{7.89}&\textbf{27.12}&\textbf{23.73}&\textbf{32.25}&\textbf{55.25}\\
         \hline
    \end{tabular}
    \label{tab:rewards}
\end{table}

\subsection{Limitations}
We acknowledge several limitations of our approach. First, the proposed reasoning supervision relies on the availability of region-level annotations from Chest ImaGenome to construct structured reasoning traces. Although these annotations enable explicit verification of the reasoning process, they are currently available only for chest radiographs, limiting the direct applicability of our framework to other imaging modalities or anatomical regions where comparable annotations do not exist. Second, RL substantially increases computational cost compared with SFT, requiring multiple sampled rollouts for every training example. This additional training overhead may limit the practicality of the approach for larger foundation models or resource-constrained environments. Finally, although our reward design encourages visual grounding, RL may still optimize for the reward function rather than genuine visual understanding. As with other RL trained VLMs, the learned policy may exploit imperfections in the reward design, and additional evaluation on adversarial settings is needed to fully assess robustness.

\section{Conclusion}
We propose an RL methodology that verifies both the model's reasoning process and its final report using spatial and factual rewards. Spatial rewards are conditioned on the model producing output in the correct structural format and accurate coordinates, while factual rewards are tied to the correct detection of specific anatomical regions. Our results demonstrate that verifying both the reasoning process and the final report during the RL stage leads to substantially better performance than verifying either component in isolation. Furthermore, we find that effective reasoning verification requires joint tracking of predicted bounding-box coordinates and their associated regional findings; without this joint supervision, the model tends to produce ``shortcut'' behavior, generating identical findings across different regions. This highlights the critical importance of enforcing both spatial grounding and factual consistency when training medical vision–language models. Our approach results in models that are more clinically faithful and more interpretable. The structured think-traces make the model behavior more auditable for clinicians.

\bibliography{aaai2027}

@misc{bannur2024maira2groundedradiologyreport,
      title={MAIRA-2: Grounded Radiology Report Generation}, 
      author={Shruthi Bannur and Kenza Bouzid and Daniel C. Castro and Anton Schwaighofer and Anja Thieme and Sam Bond-Taylor and Maximilian Ilse and Fernando Pérez-García and Valentina Salvatelli and Harshita Sharma and Felix Meissen and Mercy Ranjit and Shaury Srivastav and Julia Gong and Noel C. F. Codella and Fabian Falck and Ozan Oktay and Matthew P. Lungren and Maria Teodora Wetscherek and Javier Alvarez-Valle and Stephanie L. Hyland},
      year={2024},
      eprint={2406.04449},
      archivePrefix={arXiv},
      primaryClass={cs.CL},
      url={https://arxiv.org/abs/2406.04449}, 
}

@inproceedings{pan2025medvlm,
  title={MedVLM-R1: Incentivizing Medical Reasoning Capability of Vision-Language Models (VLMs) via Reinforcement Learning},
  author={Pan, Jiazhen and Liu, Che and Wu, Junde and Liu, Fenglin and Zhu, Jiayuan and Li, Hongwei Bran and Chen, Chen and Ouyang, Cheng and Rueckert, Daniel},
  booktitle={International Conference on Medical Image Computing and Computer-Assisted Intervention},
  pages={337--347},
  year={2025},
  organization={Springer}
}

@inproceedings{xu2025medground,
  title={MedGround-R1: Advancing Medical Image Grounding via Spatial-Semantic Rewarded Group Relative Policy Optimization},
  author={Xu, Huihui and Nie, Yuanpeng and Wang, Hualiang and Chen, Ying and Li, Wei and Ning, Junzhi and Liu, Lihao and Wang, Hongqiu and Zhu, Lei and Liu, Jiyao and others},
  booktitle={International Conference on Medical Image Computing and Computer-Assisted Intervention},
  pages={391--401},
  year={2025},
  organization={Springer}
}

@article{johnson2019mimic,
  title={MIMIC-CXR, A De-Identified Publicly Available Database of Chest Radiographs with Free-Text Reports},
  author={Johnson, Alistair EW and Pollard, Tom J and Berkowitz, Seth J and Greenbaum, Nathaniel R and Lungren, Matthew P and Deng, Chih-ying and Mark, Roger G and Horng, Steven},
  journal={Scientific data},
  volume={6},
  number={1},
  pages={317},
  year={2019},
  publisher={Nature Publishing Group UK London}
}

@article{wu2021chest,
  title={Chest Imagenome Dataset for Clinical Reasoning},
  author={Wu, Joy T and Agu, Nkechinyere N and Lourentzou, Ismini and Sharma, Arjun and Paguio, Joseph A and Yao, Jasper S and Dee, Edward C and Mitchell, William and Kashyap, Satyananda and Giovannini, Andrea and others},
  journal={arXiv preprint arXiv:2108.00316},
  year={2021}
}

@misc{bai2025qwen3vltechnicalreport,
      title={Qwen3-VL Technical Report}, 
      author={Shuai Bai and Yuxuan Cai and Ruizhe Chen and Keqin Chen and Xionghui Chen and Zesen Cheng and Lianghao Deng and Wei Ding and Chang Gao and Chunjiang Ge and Wenbin Ge and Zhifang Guo and Qidong Huang and Jie Huang and Fei Huang and Binyuan Hui and Shutong Jiang and Zhaohai Li and Mingsheng Li and Mei Li and Kaixin Li and Zicheng Lin and Junyang Lin and Xuejing Liu and Jiawei Liu and Chenglong Liu and Yang Liu and Dayiheng Liu and Shixuan Liu and Dunjie Lu and Ruilin Luo and Chenxu Lv and Rui Men and Lingchen Meng and Xuancheng Ren and Xingzhang Ren and Sibo Song and Yuchong Sun and Jun Tang and Jianhong Tu and Jianqiang Wan and Peng Wang and Pengfei Wang and Qiuyue Wang and Yuxuan Wang and Tianbao Xie and Yiheng Xu and Haiyang Xu and Jin Xu and Zhibo Yang and Mingkun Yang and Jianxin Yang and An Yang and Bowen Yu and Fei Zhang and Hang Zhang and Xi Zhang and Bo Zheng and Humen Zhong and Jingren Zhou and Fan Zhou and Jing Zhou and Yuanzhi Zhu and Ke Zhu},
      year={2025},
      eprint={2511.21631},
      archivePrefix={arXiv},
      primaryClass={cs.CV},
      url={https://arxiv.org/abs/2511.21631}, 
}

@dataset{wu2021chestimagenome,
  author       = {Wu, Jing and Agu, Nestor and Lourentzou, Ismini and Sharma, Abhishek and Paguio, Joshua and Yao, James S. and Dee, Erwin C. and Mitchell, William and Kashyap, Saurabh and Giovannini, Alessandro and Celi, Leo Anthony and Syeda-Mahmood, Tanveer and Moradi, Mehdi},
  title        = {{Chest ImaGenome Dataset (version 1.0.0)}},
  year         = {2021},
  publisher    = {PhysioNet},
  note         = {RRID:SCR\_007345},
  doi          = {10.13026/wv01-y230},
  url          = {https://doi.org/10.13026/wv01-y230}
}

@article{johnson2019mimiccxr,
  author       = {Johnson, Alistair E. W. and Pollard, Tom J. and Berkowitz, Seth J. and Greenbaum, Nathaniel R. and Lungren, Matthew P. and Deng, Chih-ying and Mark, Roger G. and Horng, Steven},
  title        = {{MIMIC-CXR, A De-identified Publicly Available Database of Chest Radiographs with Free-text Reports}},
  journal      = {Scientific Data},
  year         = {2019},
  volume       = {6},
  number       = {317},
  doi          = {10.1038/s41597-019-0322-0},
  url          = {https://doi.org/10.1038/s41597-019-0322-0},
  publisher    = {Nature Publishing Group}
}

@article{liu2026scaling,
  title={Scaling Medical Imaging Report Generation with Multimodal Reinforcement Learning},
  author={Liu, Qianchu and Zhang, Sheng and Qin, Guanghui and Gu, Yu and Jin, Ying and Preston, Sam and Xu, Yanbo and Kiblawi, Sid and Yim, Wen-wai and Ossowski, Tim and others},
  journal={arXiv preprint arXiv:2601.17151},
  year={2026}
}

@inproceedings{Liu_2025, series={MM ’25},
   title={GEMeX-RMCoT: An Enhanced Med-VQA Dataset for Region-Aware Multimodal Chain-of-Thought Reasoning},
   url={http://dx.doi.org/10.1145/3746027.3758277},
   DOI={10.1145/3746027.3758277},
   booktitle={Proceedings of the 33rd ACM International Conference on Multimedia},
   publisher={ACM},
   author={Liu, Bo and Zhao, Xiangyu and He, Along and Chen, Yidi and Fu, Huazhu and Wu, Xiao-Ming},
   year={2025},
   month=oct, pages={13213–13220},
   collection={MM ’25} }

@article{gundersen2025enhancing,
  title={Enhancing Radiology Report Generation and Visual Grounding using Reinforcement Learning},
  author={Gundersen, Benjamin and Deperrois, Nicolas and Ruiperez-Campillo, Samuel and Sutter, Thomas M and Vogt, Julia E and Moor, Michael and Nooralahzadeh, Farhad and Krauthammer, Michael},
  journal={arXiv preprint arXiv: 2512.10691},
  year={2025}
}

@article{shao2024deepseekmath,
  title={Deepseekmath: Pushing The Limits of Mathematical Reasoning in Open Language Models},
  author={Shao, Zhihong and Wang, Peiyi and Zhu, Qihao and Xu, Runxin and Song, Junxiao and Bi, Xiao and Zhang, Haowei and Zhang, Mingchuan and Li, YK and Wu, Yang and others},
  journal={arXiv preprint arXiv:2402.03300},
  year={2024}
}

@article{guo2025deepseek,
  title={DeepSeek-R1 Incentivizes Reasoning in LLMs Through Reinforcement Learning},
  author={Guo, Daya and Yang, Dejian and Zhang, Haowei and Song, Junxiao and Wang, Peiyi and Zhu, Qihao and Xu, Runxin and Zhang, Ruoyu and Ma, Shirong and Bi, Xiao and others},
  journal={Nature},
  volume={645},
  number={8081},
  pages={633--638},
  year={2025},
  publisher={Nature Publishing Group UK London}
}

@article{yang2025qwen3,
  title={Qwen3 Technical Report},
  author={Yang, An and Li, Anfeng and Yang, Baosong and Zhang, Beichen and Hui, Binyuan and Zheng, Bo and Yu, Bowen and Gao, Chang and Huang, Chengen and Lv, Chenxu and others},
  journal={arXiv preprint arXiv:2505.09388},
  year={2025}
}

@article{yu2025dapo,
  title={DAPO: An Open-Source LLM Reinforcement Learning System at Scale},
  author={Yu, Qiying and Zhang, Zheng and Zhu, Ruofei and Yuan, Yufeng and Zuo, Xiaochen and Yue, Yu and Dai, Weinan and Fan, Tiantian and Liu, Gaohong and Liu, Lingjun and others},
  journal={arXiv preprint arXiv:2503.14476},
  year={2025}
}

@article{lin2025foundation,
  title={A Foundation Model for Chest X-ray Interpretation with Grounded Reasoning via Online Reinforcement Learning},
  author={Lin, Qika and Zhu, Yifan and Pu, Bin and Huang, Ling and Luo, Haoran and Ma, Jingying and Peng, Zhen and Zhao, Tianzhe and Xu, Fangzhi and Zhang, Jian and others},
  journal={arXiv preprint arXiv:2509.03906},
  year={2025}
}

@article{sellergren2025medgemma,
  title={MedGemma Technical Report},
  author={Sellergren, Andrew and Kazemzadeh, Sahar and Jaroensri, Tiam and Kiraly, Atilla and Traverse, Madeleine and Kohlberger, Timo and Xu, Shawn and Jamil, Fayaz and Hughes, C{\'\i}an and Lau, Charles and others},
  journal={arXiv preprint arXiv:2507.05201},
  year={2025}
}

@inproceedings{lin2004rouge,
  title={Rouge: A Package for Automatic Evaluation of Summaries},
  author={Lin, Chin-Yew},
  booktitle={Text summarization branches out},
  pages={74--81},
  year={2004}
}

@inproceedings{banerjee2005meteor,
  title={METEOR: An automatic metric for MT evaluation with improved correlation with human judgments},
  author={Banerjee, Satanjeev and Lavie, Alon},
  booktitle={Proceedings of the acl workshop on intrinsic and extrinsic evaluation measures for machine translation and/or summarization},
  pages={65--72},
  year={2005}
}

@article{jain2021radgraph,
  title={Radgraph: Extracting clinical entities and relations from radiology reports (2021)},
  author={Jain, Saahil and Agrawal, Ashwin and Saporta, Adriel and Truong, SQ and Duong, Du Nguyen and Bui, Tan and Chambon, Pierre and Zhang, Yuhao and Lungren, Matthew P and Ng, Andrew Y and others},
  journal={arXiv preprint arXiv:2106.14463},
  year={2021},
  publisher={URL}
}

@inproceedings{papineni-etal-2002-bleu,
    title = "{B}leu: a Method for Automatic Evaluation of Machine Translation",
    author = "Papineni, Kishore  and
      Roukos, Salim  and
      Ward, Todd  and
      Zhu, Wei-Jing",
    editor = "Isabelle, Pierre  and
      Charniak, Eugene  and
      Lin, Dekang",
    booktitle = "Proceedings of the 40th Annual Meeting of the Association for Computational Linguistics",
    month = jul,
    year = "2002",
    address = "Philadelphia, Pennsylvania, USA",
    publisher = "Association for Computational Linguistics",
    url = "https://aclanthology.org/P02-1040/",
    doi = "10.3115/1073083.1073135",
    pages = "311--318"
}

@article{bannur2024maira,
  title={Maira-2: Grounded radiology report generation},
  author={Bannur, Shruthi and Bouzid, Kenza and Castro, Daniel C and Schwaighofer, Anton and Thieme, Anja and Bond-Taylor, Sam and Ilse, Maximilian and P{\'e}rez-Garc{\'\i}a, Fernando and Salvatelli, Valentina and Sharma, Harshita and others},
  journal={arXiv preprint arXiv:2406.04449},
  year={2024}
}

@article{zhang2019bertscore,
  title={Bertscore: Evaluating text generation with bert},
  author={Zhang, Tianyi and Kishore, Varsha and Wu, Felix and Weinberger, Kilian Q and Artzi, Yoav},
  journal={arXiv preprint arXiv:1904.09675},
  year={2019}
}

@article{chambon2024chexpert,
  title={Chexpert plus: Augmenting a large chest x-ray dataset with text radiology reports, patient demographics and additional image formats},
  author={Chambon, Pierre and Delbrouck, Jean-Benoit and Sounack, Thomas and Huang, Shih-Cheng and Chen, Zhihong and Varma, Maya and Truong, Steven QH and Chuong, Chu The and Langlotz, Curtis P},
  journal={arXiv preprint arXiv:2405.19538},
  year={2024}
}

@article{li2023llava,
  title={Llava-med: Training a large language-and-vision assistant for biomedicine in one day},
  author={Li, Chunyuan and Wong, Cliff and Zhang, Sheng and Usuyama, Naoto and Liu, Haotian and Yang, Jianwei and Naumann, Tristan and Poon, Hoifung and Gao, Jianfeng},
  journal={Advances in Neural Information Processing Systems},
  volume={36},
  pages={28541--28564},
  year={2023}
}

@article{lai2026med,
  title={Med-r1: Reinforcement learning for generalizable medical reasoning in vision-language models},
  author={Lai, Yuxiang and Zhong, Jike and Li, Ming and Zhao, Shitian and Li, Yuheng and Psounis, Konstantinos and Yang, Xiaofeng},
  journal={IEEE Transactions on Medical Imaging},
  year={2026},
  publisher={IEEE}
}

@article{fan2025chestx,
  title={Chestx-reasoner: Advancing radiology foundation models with reasoning through step-by-step verification},
  author={Fan, Ziqing and Liang, Cheng and Wu, Chaoyi and Zhang, Ya and Wang, Yanfeng and Xie, Weidi},
  journal={arXiv preprint arXiv:2504.20930},
  year={2025}
}

@article{li2025think,
  title={Think or not think: A study of explicit thinking in rule-based visual reinforcement fine-tuning},
  author={Li, Ming and Zhong, Jike and Zhao, Shitian and Lai, Yuxiang and Zhang, Haoquan and Zhu, Wang Bill and Zhang, Kaipeng},
  journal={arXiv preprint arXiv:2503.16188},
  year={2025}
}

@article{huang2025vision,
  title={Vision-r1: Incentivizing reasoning capability in multimodal large language models},
  author={Huang, Wenxuan and Jia, Bohan and Zhai, Zijie and Cao, Shaosheng and Ye, Zheyu and Zhao, Fei and Xu, Zhe and Hu, Yao and Lin, Shaohui},
  journal={arXiv preprint arXiv:2503.06749},
  year={2025}
}

@inproceedings{liu2025visual,
  title={Visual-rft: Visual reinforcement fine-tuning},
  author={Liu, Ziyu and Sun, Zeyi and Zang, Yuhang and Dong, Xiaoyi and Cao, Yuhang and Duan, Haodong and Lin, Dahua and Wang, Jiaqi},
  booktitle={Proceedings of the IEEE/CVF International Conference on Computer Vision},
  pages={2034--2044},
  year={2025}
}

@article{shen2025vlm,
  title={Vlm-r1: A stable and generalizable r1-style large vision-language model},
  author={Shen, Haozhan and Liu, Peng and Li, Jingcheng and Fang, Chunxin and Ma, Yibo and Liao, Jiajia and Shen, Qiaoli and Zhang, Zilun and Zhao, Kangjia and Zhang, Qianqian and others},
  journal={arXiv preprint arXiv:2504.07615},
  year={2025}
}

@inproceedings{xu2025radeval,
  title={RadEval: A framework for radiology text evaluation},
  author={Xu, Justin and Zhang, Xi and Abderezaei, Javid and Bauml, Julie and Boodoo, Roger and Haghighi, Fatemeh and Ganjizadeh, Ali and Brattain, Eric and Van Veen, Dave and Meng, Zaiqiao and others},
  booktitle={Proceedings of the 2025 Conference on Empirical Methods in Natural Language Processing: System Demonstrations},
  pages={546--557},
  year={2025}
}

@article{irede2026medical,
  title={Medical imaging: a critical review on X-ray imaging for the detection of infection},
  author={Irede, Egwonor Loveth and Aworinde, Omowunmi Rebecca and Lekan, Ogunnaike Korede and Amienghemhen, Osemudiamhen D and Okonkwo, Tochukwu Perpetua and Onivefu, Asishana Paul and Ifijen, Ikhazuagbe H},
  journal={Biomedical Materials \& Devices},
  volume={4},
  number={1},
  pages={1--45},
  year={2026},
  publisher={Springer}
}

@article{tanno2025collaboration,
  title={Collaboration between clinicians and vision--language models in radiology report generation},
  author={Tanno, Ryutaro and Barrett, David GT and Sellergren, Andrew and Ghaisas, Sumedh and Dathathri, Sumanth and See, Abigail and Welbl, Johannes and Lau, Charles and Tu, Tao and Azizi, Shekoofeh and others},
  journal={Nature Medicine},
  volume={31},
  number={2},
  pages={599--608},
  year={2025},
  publisher={Nature Publishing Group US New York}
}

@inproceedings{yang2025mitigating,
  title={Mitigating hallucinations in large vision-language models via dpo: On-policy data hold the key},
  author={Yang, Zhihe and Luo, Xufang and Han, Dongqi and Xu, Yunjian and Li, Dongsheng},
  booktitle={Proceedings of the IEEE/CVF Conference on Computer Vision and Pattern Recognition},
  pages={10610--10620},
  year={2025}
}

@inproceedings{heiman2025factchexcker,
  title={Factchexcker: Mitigating measurement hallucinations in chest x-ray report generation models},
  author={Heiman, Alice and Zhang, Xiaoman and Chen, Emma and Kim, Sung Eun and Rajpurkar, Pranav},
  booktitle={Proceedings of the Computer Vision and Pattern Recognition Conference},
  pages={30787--30796},
  year={2025}
}

@article{bai2024hallucination,
  title={Hallucination of multimodal large language models: A survey},
  author={Bai, Zechen and Wang, Pichao and Xiao, Tianjun and He, Tong and Han, Zongbo and Zhang, Zheng and Shou, Mike Zheng},
  journal={arXiv preprint arXiv:2404.18930},
  year={2024}
}

@article{ouyang2022training,
  title={Training language models to follow instructions with human feedback},
  author={Ouyang, Long and Wu, Jeffrey and Jiang, Xu and Almeida, Diogo and Wainwright, Carroll and Mishkin, Pamela and Zhang, Chong and Agarwal, Sandhini and Slama, Katarina and Ray, Alex and others},
  journal={Advances in neural information processing systems},
  volume={35},
  pages={27730--27744},
  year={2022}
}

@misc{schulman2020kl,
  author = {Schulman, John},
  title = {Approximating {KL} Divergence},
  year = {2020},
  howpublished = {\url{http://joschu.net/blog/kl-approx.html}}
}


\end{document}